\documentclass[11pt]{article}

\usepackage{acl}

\usepackage{times}
\usepackage{latexsym}

\usepackage[T1]{fontenc}
\usepackage[utf8]{inputenc}

\usepackage{microtype}
\usepackage{adjustbox}
\usepackage{amsmath}
\usepackage{booktabs}
\usepackage{graphicx}
\usepackage{pgfplots}
\usepackage{xcolor}

\title{Mitigating Hallucinations and Off-target Machine Translation with Source-Contrastive and Language-Contrastive Decoding}

\author{Rico Sennrich$^{1,2}$ ~~Jannis Vamvas$^{1}$ ~~Alireza Mohammadshahi$^{1,3}$
\vspace{0.1cm}\\ 
 $^1$University of Zurich ~~~~ $^2$University of Edinburgh  ~~~~~ $^3$EPFL\\
 \texttt{\{sennrich,vamvas\}@cl.uzh.ch}\\\texttt{alireza.mohammadshahi@epfl.ch}
}

\begin{document}
\maketitle
\begin{abstract}
Hallucinations and off-target translation remain unsolved problems in MT, especially for low-resource languages and massively multilingual models.
In this paper, we introduce two related methods to mitigate these failure cases with a modified decoding objective, without either requiring retraining or external models.
In source-contrastive decoding, we search for a translation that is probable given the correct input, but improbable given a random input segment.
In language-contrastive decoding, we search for a translation that is probable, but improbable given the wrong language indicator token.
Experiments on the massively multilingual models M2M-100 (418M) and SMaLL-100 show that these methods suppress hallucinations and off-target translations, reducing the number of translations with segment-level chrF2 below~10 by 67-83\% on average, and the number of translations with oscillatory hallucinations by 75-92\% on average, across 57 tested translation directions.
In a proof of concept on out-of-English translation, we also show that we can suppress off-target translations with large language models.
We release our source code.\footnote{\url{https://github.com/ZurichNLP/ContraDecode}}

\end{abstract}

\section{Introduction}

Hallucinations are a long-standing well-known problem in machine translation~(MT)~\cite{koehn-knowles-2017-six} and natural language generation \cite{jietal2023}.
While there has been extensive research on their identification and mitigation \cite[][among others]{lee2019hallucinations,raunak-etal-2021-curious,mohammadshahi-etal-2022-compressed,guerreiro2023hallucinations,dale-etal-2023-detecting}, they still persist as an issue, especially in low-resource settings.

\begin{figure}[t!]
\centering
\includegraphics[width=\columnwidth]{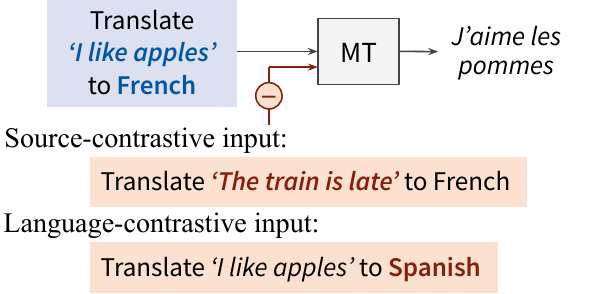}
\caption{
Our decoding objective yields a translation that is probable given the actual input, but improbable given a source-contrastive or language-contrastive input.
}
\label{fig:figure-1}
\end{figure}

Contrastive conditioning has previously been used for analysing specific translation errors such as disambiguation errors and undertranslation~\cite{vamvas-sennrich-2021-contrastive,vamvas-sennrich-2022-little}. The main idea is that translations that are equally or more probable given some corrupted source than the true source
are likely to be erroneous with respect to the corrupted span.
We can apply the same intuition to hallucinations and translations into the wrong language, so called  off-target translations: if hallucinations are detached from the source, they should have a similar probability given the true source and given a random other source.
A translation in the wrong language should have a similar or higher probability if that language is marked as desired.

Inspired by this, we design decoding objectives that do not just search for the most probable translation, but search for a translation that maximizes the probability given the true input, but minimizes the probability given one or several contrastive inputs.

This paper makes the following contributions:

\begin{itemize}
\item We introduce contrastive decoding objectives to address two problems often observed in MT: hallucinations and off-target translations.
\item By evaluating two massively multilingual MT models, M2M-100 (418M) and SMaLL-100, across 57 mostly low-resource translation directions, we show improvements in chrF2 by 1.3--1.7 points, and reduce the number of translations with chrF2 below 10 by 67-83\%.
\item Finally, we provide a proof of concept for applying our approach to LLM-based translation, where off-target issues are common.
\end{itemize}

\section{Method}

To suppress hallucinations, we pair each input~$X$ with a randomly selected input segment $X'$.\footnote{In practice, by shuffling segments of the input document.}
Rather than finding a translation that maximizes $p(Y|X)$, we search for one that both maximizes $p(Y|X)$ and minimizes $p(Y|X')$. We add a hyperparameter $\lambda$ to control the strength of this contrastive penalty, yielding Eq.~\ref{eq1}.
\begin{align}
s(Y,X) = \sum_{i=1}^{|Y|}- \log \biggl(&p(y_i| y_{<i}, X) \nonumber\\
-\lambda &p(y_i| y_{<i}, X')\biggl)
\label{eq1}
\end{align}
We denote this \textbf{source-contrastive decoding}.

Off-target translations are a common failure mode in multilingual MT systems \cite{DBLP:journals/corr/abs-1903-07091}.
They have been linked to the predominance of English in the training of multilingual systems \cite{rios-etal-2020-subword}.
Production of text in the source language, often a copy of the input, is connected to the occurrence of copying in the training data, and the high probability of continuing to copy once a copy has been started \cite{ott2018analyzing}.

The majority of multilingual MT systems use special tokens to indicate the target language, following \citet{johnson-etal-2017-googles}.\footnote{The indicator token can be in the source (SMaLL-100), or output-initial and force-decoded (M2M-100).}
To penalize output in the wrong language, we can add contrastive inputs that only vary the language indicator token.

Let $l_{y}$ be the target language.
We replace its indicator token with contrastive variants $l_{y'} \in L_c$ for languages we wish to suppress.
Based on the predominant off-target languages in multilingual MT \cite{DBLP:journals/corr/abs-1903-07091}, our set of contrastive languages $L_c$ consists of English\footnote{Unless English is the target language.} and the respective source language.
This results in Eq.~\ref{eq2}.
\begin{align}
s(Y,X) = \sum_{i=1}^{|Y|}- \log \biggl(&p(y_i| y_{<i}, X,l_{y}) \nonumber\\
-\sum_{l_{y'} \in L_c} \lambda &p(y_i| y_{<i}, X, l_{y'})\biggl)
\label{eq2}
\end{align}
We refer to decoding with contrastive translation directions as \textbf{language-contrastive decoding}.
We can combine source-contrastive and language-contrastive decoding by summing all contrastive variants, and refer to the weights as $\lambda_{\text{src}}$ and $\lambda_{\text{lang}}$.

\section{Evaluation}

\begin{table*}
\centering
\footnotesize
\begin{tabular}{lcccccccc}
\toprule
& \multicolumn{4}{c}{chrF2} & \multicolumn{4}{c}{spBLEU}\\
& HLMT & X-branch & high-res & all & HLMT & X-branch & high-res & all \\
\cmidrule(r){2-5} \cmidrule(l){6-9}
&\multicolumn{8}{c}{M2M-100}\\
baseline & 46.4 & 28.8 & 61.3 & 39.0 & 22.0 & 8.3 & 37.2 & 16.4\\
$C_{src}$ & 46.7 & 31.4 & 60.8 & 40.3 & 21.6 & 9.1 & 36.4 & 16.6\\
$C_{src+lang}$ & 46.8 & 32.1 & 60.7 & 40.7 & 21.5 & 9.3 & 36.1 & 16.6\\
\midrule
&\multicolumn{8}{c}{SMaLL-100}\\
baseline & 48.3 & 32.0 & 62.5 & 41.4 & 23.5 & 10.2 & 38.7 & 18.1\\
$C_{src}$ & 48.5 & 34.2 & 62.1 & 42.5 & 23.2 & 11.1 & 37.9 & 18.4\\
$C_{src+lang}$& 48.7 & 34.6 & 62.0 & 42.7 & 23.3 & 11.2 & 37.6 & 18.4\\
\bottomrule
\end{tabular}
\caption{Automatic evaluation results. Averages over different sets of translation directions.}
\label{main-m2m}
\end{table*}

\subsection{Data and Models}

We perform experiments with two massively multilingual MT models: M2M-100 (418M)~\cite{fan2020englishcentric}, and SMaLL-100~\cite{mohammadshahi-etal-2022-small}, a distilled version of M2M-100 (12B).

We use beam size 5.
We perform minimal hyper-parameter tuning on the ps-ast translation direction with M2M-100 and set $\lambda_{\text{src}}$ to $0.7$.\footnote{We exclude ps-ast from average results reported.}
Since only a small number of directions suffer from off-target outputs, we do not tune $\lambda_{\text{lang}}$, setting it to $0.1$.

We test on three sets of translation directions:
\begin{itemize}
\item the 25 non-English-centric directions used by \citet{guerreiro2023hallucinations} (\textbf{HLMT}). These are af-zu, ar-fr, be-ru, cs-sk, de-hr, de-hu, el-tr, fr-sw, hi-bn, hi-mr, hr-cs, hr-hu, hr-sk, hr-sr, it-de, it-fr, nl-de, nl-fr, ro-de, ro-hu, ro-hy, ro-ru, ro-tr, ro-uk, uk-ru.\footnote{See Appendix\ \ref{sec:appendix:langs} for full language names.}
\item 29 directions between 5 low-resource languages from different branches of Indo-European, plus Zulu from the Atlantic-Congo family (\textbf{X-branch}): af, ast, hr, ps, ur, zu.
\item 4 high-resource translation directions: en-de, de-en, en-fr, fr-en (\textbf{high-res}).
\end{itemize}
We also report results for the union of the sets (\textbf{all}).

We evaluate with spBLEU~\cite{goyal-etal-2022-flores}
and chrF2~\cite{popovic-2015-chrf} using sacreBLEU \cite{post-2018-call}\footnote{BLEU|\#:1|c:mixed|e:no|tok:flores101|s:exp|v:2.3.1\\
chrF2|\#:1|c:mixed|e:yes|nc:6|nw:0|s:no|v:2.3.1}
on the Flores-101 devtest set \cite{goyal-etal-2022-flores}.
We use OpenLID \cite{burchell-etal-2023-open} for language identification to measure off-target rates.
To quantify the number of hallucinations, we employ a rough approximation following \citet{lee2019hallucinations,muller-sennrich-2021-understanding}, counting the proportion of segments with chrF2 $<10$.\footnote{\citet{muller-sennrich-2021-understanding} report a threshold of 1, but this is a typo (personal communication). This method does not distinguish between hallucinations and off-target translations.}
Another automatic metric specific for oscillatory hallucinations is top n-gram (TNG)~\cite{guerreiro-etal-2023-looking,raunak2022salted,raunak-etal-2021-curious},
which measures the number of sentences whose top repeating $n$-gram is more frequent than the top repeated source $n$-gram by at least $t$.\footnote{We follow \citet{guerreiro-etal-2023-looking} and use $n=4$ and $t=2$.}

\subsection{Results}

We report results using source-contrastive decoding ($C_{src}$), and combining source-contrastive and language-contrastive decoding ($C_{src+lang}$) in Table \ref{main-m2m}.\footnote{See Appendix~\ref{sec:appendix} for full results.}
Across 57 translation directions, chrF2 improves by 1.3 (M2M-100) and 1.1 (SMaLL-100) points with source-contrastive decoding.
Language-contrastive decoding brings additional gains of 0.4 (M2M-100) and 0.2 (SMaLL-100) points.

Improvements are more modest when measured with spBLEU (0.2 on M2M-100; 0.3 on SMaLL-100).
We notice that hallucinations tend to be overlong, and can perversely improve BLEU by reducing the brevity penalty.
We thus consider chrF2, which pairs precision with recall instead of a simplistic brevity penalty, to be our primary metric.

Off-target translations are relatively rare for the translation directions tested, especially for SMaLL-100 (see Table\ \ref{offtarget}).
With M2M-100, the highest proportion of English outputs in the baseline was detected for af-zu (9.1\%), the highest percentage of outputs in the source language for hr-sr (4.2\%)\footnote{This number may be an overestimate due to the close relationship between Serbian and Croatian, and the consequent difficulty of doing reliable language identification.}.
These are also among the translation directions that benefit the most from language-contrastive decoding: chrF2 increases by 2.3 for hr-sr\footnote{This improvement is somewhat coincidental because both Latin and Cyrillic are accepted for Serbian, but Flores-101 has Cyrillic references. Penalizing output in Croatian, which uses the Latin alphabet, indirectly rewards output in Cyrillic.}, and by 2 for af-zu.
However, we observe the largest increase in chrF2 (3.2) for ast-zu, a direction where source-contrastive decoding increases off-target outputs, and where the English output rate goes from 5.5\% (baseline) to 9.9\% ($C_{src}$) to 2.7\% ($C_{src+lang}$).

\begin{table}
\centering
\footnotesize
\begin{tabular}{lcccc}
\toprule
& \multicolumn{2}{c}{M2M-100} & \multicolumn{2}{c}{SMaLL-100}\\
& EN & SRC & EN & SRC\\
\cmidrule(r){2-3} \cmidrule(l){4-5}
baseline & 260 & 55 & 54 & 63\\
$C_{src}$ & 375 & 47 & 78 & 70\\
$C_{src+lang}$  & \phantom{0}88 & 28 & 16 & 21 \\
\bottomrule
\end{tabular}
\caption{Number of off-target outputs (out of 57684), in English (EN) or the source language (SRC).}
\label{offtarget}
\end{table}

\begin{table}
\centering
\footnotesize
\begin{tabular}{lcccc}
\toprule
& HLMT & X-branch & high-res & all \\
\cmidrule(r){2-5}
& \multicolumn{4}{c}{M2M-100}\\
baseline & 2.1\% & 13.0\% & 0.0\% & 7.3\% \\
$C_{src}$ & 1.0\% & \phantom{0}4.1\% & 0.0\% & 2.4\%\\
$C_{src+lang}$ & 0.5\% & \phantom{0}2.0\% & 0.0\% & 1.2\%\\
\midrule
& \multicolumn{4}{c}{SMaLL-100}\\
baseline & 1.3\% & 10.6\% & 0.0\% & 5.6\% \\
$C_{src}$ & 0.8\% & \phantom{0}4.3\% & 0.0\% & 2.5\% \\
$C_{src+lang}$ & 0.4\% & \phantom{0}3.4\% & 0.0\% & 1.8\% \\
\bottomrule
\end{tabular}
\caption{Proportion of translations with chrF2 $< 10$.}
\label{hall-results}
\end{table}

\begin{table}
\centering
\footnotesize
\begin{tabular}{lcccc}
\toprule
& HLMT & X-branch & high-res & all \\
\cmidrule(r){2-5}
& \multicolumn{4}{c}{M2M-100}\\
baseline & 2.4\% & 16.9\% & 0.0\% & 9.3\% \\
$C_{src}$ & 0.3\% & \phantom{0}3.7\% & 0.0\% & 2.0\%\\
$C_{src+lang}$ & 0.1\% & \phantom{0}1.3\% & 0.0\% & 0.7\%\\
\midrule
& \multicolumn{4}{c}{SMaLL-100}\\
baseline & 0.7\% & 11.2\% & 0.0\% & 5.9\% \\
$C_{src}$ & 0.1\% & \phantom{0}3.9\% & 0.0\% & 2.0\%\\
$C_{src+lang}$ & 0.1\% & \phantom{0}2.9\% & 0.0\% & 1.5\%\\
\bottomrule
\end{tabular}
\caption{Proportion of translations with oscillatory hallucinations according to TNG.}
\label{tng-results}
\end{table}

The proportion of translations with chrF2 below~10 is shown in Table\ \ref{hall-results}.
We observe large reductions in the number of defect translations, with a reduction from 7.3\% to 1.2\% (-83\%) for M2M-100, and from 5.6\% to 1.8\% (-67\%) for SMaLL-100.
When focusing on oscillatory hallucinations according to TNG in Table\ \ref{tng-results}, the improvement is even more pronounced,
with a reduction from 9.3\% to 0.7\% (-92\%) for M2M-100, and from 5.9\% to 1.5\% \mbox{(-75\%)} for SMaLL-100.

\section{Ablation Studies}

The fact that we pick contrastive inputs from the test sets at random raises a few questions about this approximation.
We repeated the translation with M2M-100 across all 57 translation directions 3 times and find that the standard deviation is minimal (0.0107 for chrF2).
Using a single random input as a contrastive variant is a heavy approximation, but our ablation study in Table\ \ref{ablation-m2m} shows that this yields the majority of the performance gains, and using up to 3 inputs as contrastive examples\footnote{we divide $\lambda_{\text{src}}$ by the number of contrastive inputs.} only yields an additional 0.1 point improvement in chrF2.

\begin{table}[hbt!]
\centering
\footnotesize
\begin{tabular}{lccc}
\toprule
 & chrF2 & spBLEU\\
\cmidrule(lr){2-3}
baseline & 38.97 & 16.40 \\
$C_{src}$ (1) & 40.31 & 16.60 \\
$C_{src}$ (2) & 40.39 & 16.68 \\
$C_{src}$ (3) & 40.41 & 16.67 \\
\bottomrule
\end{tabular}
\caption{Ablation results for M2M-100 with different numbers of source-contrastive inputs. Average over all languages reported.}
\label{ablation-m2m}
\end{table}

\section{Application to Large Language Models}

In this section, we demonstrate that our method can be applied to large language models~(LLM).
Previous work has achieved competitive translation quality for some directions by prompting models such as PaLM~\cite{vilar-etal-2023-prompting,pmlr-v202-garcia23a}, GPT~\cite{hendy2023good} or BLOOM~\cite{bawden-yvon-2023-investigating}.
However, LLM-based translation is still prone to hallucination and off-target translation~\cite{pmlr-v202-zhang23m,guerreiro2023hallucinations}.

Our demonstration is based on the Llama 2 model family~\cite{touvron2023llama} and specifically the instruction-tuned version (\textit{Llama Chat}), exploiting the fact that MT examples were among the data used for instruction tuning~\cite{wei2022finetuned,chung2022scaling}.
We generate translations by instructing the model to translate a segment into a given language, force-decoding the line \textit{``Sure, here's the translation:''}, and then decoding until the next line break.
The prompt we used is detailed in Appendix~\ref{sec:llm-implementation-details}.

When using this simple prompting approach in the en--de direction, we find that off-target output in English is very common.
Moreover, providing a \mbox{1-shot} example in the prompt, while improving translation quality, does not prevent the off-target issue.
We thus apply language-contrastive decoding and add a contrastive prompt that instructs the model to ``translate'' into English instead of German.
The decoding objective is analogous to Eq.~\ref{eq2}.
We use 4-bit precision~\citep{pmlr-v202-dettmers23a} and greedy decoding.

\begin{figure}
\centering
\resizebox{\columnwidth}{!}{%
\begin{tikzpicture}
\begin{axis}[
    height=6.5cm,
    width=9cm,
    axis lines = left,
    xlabel = $\lambda_{\text{lang}}$,
    ylabel = {off-target rate},
    ymin=0, ymax=10,
    xtick={0,0.1,0.3,0.5,0.7,0.9},
    yticklabel={\pgfmathprintnumber\tick\%},
    legend pos=north east,
    legend cell align={left},
]
\addplot[
    color={rgb,255:red,153; green,169; blue,219},
    mark=*,
    line width=1.5pt,
    dashed,
    ]
    coordinates {
    (0,5.04) (0.1,4.35) (0.3,3.16) (0.5,2.17) (0.7,1.78) (0.9,1.88)
    };
\addlegendentry{Llama 7B 0-shot}
\addplot[
    color={rgb,255:red,153; green,169; blue,219},
    mark=*,
    line width=1.5pt,
    ]
    coordinates {
    (0,6.23) (0.1,5.04) (0.3,4.05) (0.5,2.96) (0.7,2.96) (0.9,3.06)
    };
\addlegendentry{Llama 7B 1-shot}
\addplot[
    color={rgb,255:red,0; green,40; blue,165},
    mark=*,
    dashed,
    line width=1.5pt,
    ]
    coordinates {
    (0,3.75) (0.1,3.46) (0.3,2.67) (0.5,1.88) (0.7,1.98) (0.9,2.37)
    };
\addlegendentry{Llama 13B 0-shot}
\addplot[
    color={rgb,255:red,0; green,40; blue,165},
    mark=*,
    line width=1.5pt,
    ]
    coordinates {
    (0,5.14) (0.1,4.45) (0.3,3.75) (0.5,2.47) (0.7,1.88) (0.9,1.58)
    };
\addlegendentry{Llama 13B 1-shot}
\end{axis}
\end{tikzpicture}
}
\caption{Off-target translation rate for Llama~2 Chat models when translating the English Flores-101 devtest set into German.
Language-contrastive decoding tends to reduce off-target translation as $\lambda_{\text{lang}}$ is increased.
}
\label{fig:llm-en-de}
\end{figure}
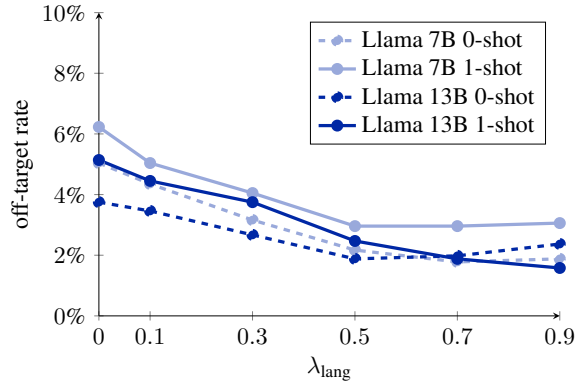

Figure~\ref{fig:llm-en-de} shows the percentage of off-target output for different $\lambda_{\text{lang}}$.
Generally, we observe that the off-target rate falls with increasing $\lambda_{\text{lang}}$, demonstrating the effectiveness of our method with LLM prompting.
English--French~(Appendix~\ref{sec:llm-results-en-fr}) has similar results.
In terms of overall translation quality, we find that language-contrastive decoding improves chrF2 and spBLEU and only becomes detrimental for $\lambda_{\text{lang}} > 0.7$ (Appendix~\ref{sec:llm-results}).

\section{Related Work}

\subsubsection*{Hallucination Detection and Reduction}

Various methods have been proposed to detect hallucinations, including identifying typical patterns in the output \cite{raunak-etal-2021-curious}, using internal information like attention patterns \cite{lee2019hallucinations} or the contribution of the source to predictions \cite{dale-etal-2023-detecting}, or measures of decoder confidence, including the output probability \cite{guerreiro-etal-2023-looking} or stability of samples under perturbation \cite{lee2019hallucinations,guerreiro-etal-2023-looking}.

Hallucination mitigation is more difficult, especially if we assume that models are already trained with best practices, and focus on training-free methods.
Several studies use external models for mitigation, e.g.\ using other translation models as a fall-back \cite{guerreiro2023hallucinations}, or sample reranking based on quality estimation~(QE) models \cite{guerreiro-etal-2023-looking}.
Our method has the advantage of not requiring external models, and we note that modern QE metrics are themselves prone to score certain hallucinations highly~\cite{freitag-etal-2022-high,yan-etal-2023-bleurt}.

Mitigation methods that do not rely on external models are typically sampling-based.
\citet{guerreiro-etal-2023-looking} report that even the translation model's own sequence probability can be used for sample reranking.
A consensus translation can be identified via sampling-based Minimum Bayes Risk decoding \cite{eikema-aziz-2020-map}, which benefits from the fact that hallucinations are dissimilar from each other \cite{muller-sennrich-2021-understanding}.

\subsubsection*{Contrastive Decoding}

Contrastive decoding is similar to contrastive learning \cite[e.g.][]{1640964,socher-etal-2014-grounded, gao-etal-2021-simcse} in that positive and negative examples are contrasted, but involves no training.

\citet{li-etal-2023-contrastive} introduce a form of contrastive decoding that contrasts the probability between different models, whereas our methods work with a single model, contrasting inputs.
\citet{su2023contrastive} introduce a contrastive search where potential output tokens are compared to previous tokens, penalizing outputs that are similar to the context and thus suppressing repetition patterns.

Source-contrastive decoding can also be seen as a variant of implicit language model (ILM) compensation, mirroring recent work by \citet{herold-etal-2023-improving}.
Our work is different in motivation in that ILM is typically used to allow the inclusion of an external LM, where we show the effectiveness of simply suppressing the ILM.
Also, we show the effectiveness of a different, simple approximation.

Finally, language-contrastive decoding bears resemblance to negative prompting, a technique used to suppress concepts in image generation.

\section{Conclusion}

This paper shows that certain failure modes of MT can be addressed by contrastive decoding objectives that use pairs or sets of inputs for the prediction.
Specific contrastive inputs address specific errors, and we introduce strategies to mitigate hallucinations and off-target translation. 

Future work could expand on our work by exploring if other MT failure modes can be mitigated with appropriate contrastive inputs, or if other forms of control can be improved.
For example, for models that use domain indicator tokens \cite{kobus-etal-2017-domain}, we could perform domain-contrastive decoding and achieve stronger domain control.
Beyond MT, we expect that source-contrastive decoding can also be useful for other tasks, e.g.\ to penalize over-generic responses in dialogue systems.

\section{Limitations}

We only tested language-contrastive decoding in multilingual models that control the target language via language indicator tokens.
It is possible to apply the same strategy to modular architectures that use language-specific components \cite{firat-etal-2016-multi,vazquez-etal-2019-multilingual,bapna-firat-2019-simple}, but its effectiveness remains to be tested.
For bilingual translation models that suffer from off-target translations, e.g.\ because of noisy training data \cite{khayrallah-koehn-2018-impact}, we would need bilingual models for other translation directions to implement language-contrastive decoding, but this sacrifices the main strength of our approach: not relying on external models.

We perform minimal hyperparameter tuning for $\lambda_{\text{src}}$, and did not tune $\lambda_{\text{lang}}$.
Using the same hyperparameters across translation directions and translation models results in performance degradations in some cases, most noticeably for high-resource translation directions.
We consider it a positive result that we obtain improvements on average with minimal hyperparameter tuning, but future work may wish to use more complex strategies to weight (or disable) contrastive variants across translation directions.

\section{Ethics Statement}

This paper introduces new decoding objectives for machine translation, and we do not foresee any harms being caused by source-contrastive or language-contrastive decoding.
More widely, we are interested in exploring novel contrastive inputs for risk mitigation, e.g.\ for model debiasing, but certain contrastive inputs could also have undesirable consequences, e.g.\ increasing model bias.

\section*{Acknowledgements}

We thank the anonymous reviewers for their comments.
This work was funded by the Swiss National Science Foundation (project MUTAMUR; no.~213976).

\bibliography{anthology,custom}
\bibliographystyle{acl_natbib}

\appendix

\vfill

\onecolumn

\section{Full Results}
\label{sec:appendix}

\begin{table*}[hbt!]
\centering
\footnotesize
\begin{adjustbox}{width=0.7\textwidth}
\begin{tabular}{lcccccc}
\toprule
direction & \multicolumn{3}{c}{chrF2} & \multicolumn{3}{c}{spBLEU}\\
& baseline & $C_{src}$ & $C_{src+lang}$ & baseline & $C_{src}$ & $C_{src+lang}$\\
\cmidrule(r){2-4} \cmidrule(l){5-7}

af-zu & 20.0 & 24.2 & 26.2 & 3.6 & 4.1 & 4.7 \\
ar-fr & 53.5 & 52.9 & 52.3 & 27.9 & 26.8 & 25.9 \\
be-ru & 42.6 & 43.8 & 43.7 & 15.8 & 16.5 & 16.5 \\
cs-sk & 55.5 & 55.1 & 55.0 & 33.7 & 33.0 & 32.8 \\
de-hr & 50.1 & 50.1 & 50.2 & 23.0 & 22.6 & 22.8 \\
de-hu & 49.1 & 48.7 & 48.8 & 23.2 & 22.3 & 22.3 \\
el-tr & 46.2 & 46.4 & 46.3 & 19.6 & 19.6 & 19.4 \\
fr-sw & 41.9 & 44.0 & 44.0 & 15.3 & 15.8 & 15.8 \\
hi-bn & 36.5 & 37.3 & 37.8 & 16.1 & 16.2 & 16.4 \\
hi-mr & 34.6 & 34.7 & 35.1 & 10.5 & 10.3 & 10.3 \\
hr-cs & 48.6 & 48.1 & 47.9 & 26.3 & 25.4 & 25.0 \\
hr-hu & 48.2 & 47.6 & 47.7 & 21.7 & 20.8 & 20.9 \\
hr-sk & 49.7 & 49.4 & 49.3 & 26.9 & 26.2 & 26.0 \\
hr-sr & 48.4 & 48.2 & 50.5 & 28.0 & 27.8 & 28.8 \\
it-de & 50.1 & 49.8 & 49.6 & 22.0 & 21.5 & 21.3 \\
it-fr & 56.8 & 56.2 & 55.9 & 32.7 & 31.7 & 30.9 \\
nl-de & 49.6 & 49.1 & 48.8 & 21.2 & 20.7 & 20.5 \\
nl-fr & 51.7 & 51.1 & 50.6 & 26.7 & 25.8 & 25.1 \\
ro-de & 52.5 & 52.3 & 52.1 & 25.0 & 24.7 & 24.3 \\
ro-hu & 49.5 & 49.1 & 48.8 & 23.5 & 22.8 & 22.6 \\
ro-hy & 24.1 & 28.7 & 29.3 & 4.7 & 6.3 & 6.4 \\
ro-ru & 48.7 & 48.4 & 48.3 & 23.6 & 23.1 & 22.8 \\
ro-tr & 50.3 & 50.4 & 50.3 & 24.2 & 24.0 & 23.7 \\
ro-uk & 48.2 & 47.9 & 47.9 & 23.8 & 23.4 & 23.4 \\
uk-ru & 53.8 & 53.4 & 53.3 & 29.9 & 29.5 & 29.3 \\
\midrule
avg (non-English-centric) & 46.4 & 46.7 & 46.8 & 22.0 & 21.6 & 21.5 \\
\midrule
\midrule
af-ast & 45.1 & 46.3 & 46.2 & 19.3 & 19.2 & 18.9 \\
af-hr & 47.6 & 47.4 & 47.4 & 20.8 & 20.3 & 20.3 \\
af-ps & 22.8 & 24.4 & 24.5 & 5.4 & 5.7 & 5.8 \\
af-ur & 35.9 & 36.4 & 36.5 & 14.0 & 14.1 & 14.1 \\
af-zu & 20.0 & 24.2 & 26.2 & 3.6 & 4.1 & 4.7 \\
ast-af & 39.6 & 43.0 & 42.9 & 14.2 & 15.8 & 15.8 \\
ast-hr & 33.7 & 41.6 & 42.7 & 11.1 & 15.8 & 16.3 \\
ast-ps & 16.6 & 21.6 & 22.4 & 2.4 & 4.7 & 4.8 \\
ast-ur & 22.2 & 31.3 & 32.0 & 6.3 & 10.7 & 10.8 \\
ast-zu & 16.0 & 21.1 & 24.3 & 2.6 & 3.3 & 3.9 \\
hr-af & 46.3 & 46.4 & 46.3 & 17.6 & 17.5 & 17.5 \\
hr-ast & 45.3 & 46.5 & 46.4 & 18.8 & 18.6 & 18.6 \\
hr-ps & 21.8 & 23.4 & 23.7 & 4.4 & 5.0 & 5.1 \\
hr-ur & 35.1 & 35.8 & 36.1 & 13.6 & 13.6 & 13.8 \\
hr-zu & 18.6 & 23.0 & 24.9 & 3.0 & 3.6 & 4.1 \\
ps-af & 34.9 & 35.5 & 36.0 & 8.3 & 8.5 & 8.7 \\
\textit{ps-ast} & \textit{32.2} & \textit{34.3} & \textit{34.2} & \textit{7.8} & \textit{9.4} & \textit{9.1} \\
ps-hr & 33.5 & 34.0 & 34.0 & 8.0 & 8.1 & 8.2 \\
ps-ur & 30.8 & 31.4 & 31.4 & 9.8 & 10.1 & 10.1 \\
ps-zu & 16.2 & 21.0 & 23.9 & 1.8 & 2.4 & 2.8 \\
ur-af & 35.3 & 36.1 & 36.6 & 9.0 & 9.1 & 9.3 \\
ur-ast & 29.7 & 33.6 & 34.1 & 7.1 & 9.1 & 9.1 \\
ur-hr & 34.2 & 35.1 & 35.4 & 8.9 & 9.1 & 9.2 \\
ur-ps & 21.2 & 22.8 & 23.5 & 4.2 & 4.8 & 4.9 \\
ur-zu & 16.0 & 19.5 & 22.2 & 1.4 & 1.7 & 2.1 \\
zu-af & 28.9 & 30.6 & 31.0 & 6.9 & 7.7 & 7.7 \\
zu-ast & 26.0 & 29.1 & 29.5 & 5.8 & 7.5 & 7.5 \\
zu-hr & 27.9 & 28.4 & 28.8 & 6.2 & 6.3 & 6.4 \\
zu-ps & 12.2 & 17.1 & 17.4 & 1.3 & 2.8 & 2.7 \\
zu-ur & 22.6 & 24.7 & 24.9 & 4.8 & 5.8 & 5.8 \\
\midrule
avg (X-branch)& 28.8 & 31.4 & 32.1 & 8.3 & 9.1 & 9.3 \\
\midrule
\midrule
de-en & 61.4 & 61.2 & 61.0 & 36.6 & 36.0 & 35.9 \\
en-de & 57.2 & 56.6 & 56.5 & 31.1 & 30.1 & 29.8 \\
en-fr & 63.8 & 63.0 & 62.9 & 42.2 & 40.9 & 40.5 \\
fr-en & 62.8 & 62.5 & 62.4 & 38.9 & 38.6 & 38.4 \\
\midrule
avg (high-res) & 61.3 & 60.8 & 60.7 & 37.2 & 36.4 & 36.1\\
\midrule
\midrule
avg (all) & 39.0 & 40.3 & 40.7 & 16.4 & 16.6 & 16.6 \\
\bottomrule
\end{tabular}
\end{adjustbox}
\caption{Full results for M2M-100. The direction ps-ast was used to tune $\lambda_{src}$ and is excluded from the averages.}
\label{all-m2m}
\end{table*}

\begin{table*}[hbt!]
\centering
\footnotesize
\begin{adjustbox}{width=0.7\textwidth}
\begin{tabular}{lcccccc}
\toprule
direction & \multicolumn{3}{c}{chrF2} & \multicolumn{3}{c}{spBLEU}\\
& baseline & $C_{src}$ & $C_{src+lang}$ & baseline & $C_{src}$ & $C_{src+lang}$\\
\cmidrule(r){2-4} \cmidrule(l){5-7}
af-zu & 26.2 & 31.4 & 31.8 & 4.4 & 6.9 & 7.0 \\  
ar-fr & 53.9 & 53.6 & 53.3 & 28.2 & 27.7 & 27.0 \\
be-ru & 45.1 & 45.2 & 45.1 & 17.3 & 17.5 & 17.3 \\
cs-sk & 55.3 & 55.1 & 55.2 & 33.0 & 32.6 & 32.8 \\
de-hr & 51.2 & 51.3 & 51.1 & 24.5 & 24.3 & 24.1 \\
de-hu & 49.7 & 49.4 & 49.5 & 23.7 & 23.1 & 23.1 \\
el-tr & 46.2 & 46.2 & 46.1 & 19.0 & 18.5 & 18.3 \\
fr-sw & 48.9 & 50.1 & 50.2 & 22.9 & 23.3 & 23.3 \\
hi-bn & 43.1 & 43.1 & 42.6 & 24.0 & 23.4 & 22.8 \\
hi-mr & 38.8 & 38.8 & 38.9 & 14.8 & 14.2 & 14.5 \\      
hr-cs & 49.3 & 48.9 & 49.0 & 26.3 & 25.7 & 26.1 \\
hr-hu & 49.2 & 49.0 & 48.8 & 22.5 & 22.2 & 22.1 \\
hr-sk & 50.8 & 50.4 & 50.4 & 27.8 & 27.2 & 27.1 \\
hr-sr & 47.3 & 47.1 & 52.6 & 28.0 & 27.7 & 30.5 \\ 
it-de & 51.0 & 51.2 & 51.1 & 23.5 & 23.5 & 23.3 \\
it-fr & 57.2 & 56.8 & 56.8 & 33.1 & 32.0 & 31.9 \\
nl-de & 50.2 & 50.2 & 50.1 & 22.1 & 22.0 & 21.8 \\
nl-fr & 52.7 & 52.2 & 52.2 & 27.8 & 26.8 & 26.7 \\
ro-de & 54.2 & 53.6 & 53.7 & 27.4 & 26.4 & 26.4 \\
ro-hu & 50.0 & 50.1 & 49.9 & 23.8 & 23.7 & 23.5 \\
ro-hy & 34.5 & 35.3 & 35.9 & 11.0 & 11.3 & 11.6 \\
ro-ru & 49.4 & 49.3 & 49.3 & 24.1 & 23.7 & 23.8 \\
ro-tr & 50.4 & 50.2 & 50.0 & 23.5 & 23.0 & 22.9 \\
ro-uk & 49.2 & 49.0 & 49.2 & 24.5 & 24.1 & 24.1 \\
uk-ru & 54.1 & 53.8 & 53.9 & 30.1 & 29.7 & 29.7 \\
\midrule
avg (non-English-centric) & 48.3 & 48.5 & 48.7 & 23.5 & 23.2 & 23.3 \\
\midrule\midrule
af-ast & 48.3 & 49.7 & 49.3 & 22.0 & 21.6 & 21.5 \\
af-hr & 50.6 & 50.6 & 50.4 & 23.5 & 23.4 & 23.3 \\
af-ps & 24.8 & 24.9 & 25.1 & 6.4 & 6.2 & 6.1 \\
af-ur & 36.3 & 36.3 & 36.7 & 13.9 & 13.8 & 14.0 \\
af-zu & 26.2 & 31.4 & 31.8 & 4.4 & 6.9 & 7.0 \\
ast-af & 49.2 & 49.4 & 49.5 & 22.8 & 22.7 & 22.7 \\
ast-hr & 47.1 & 47.9 & 47.9 & 21.1 & 21.1 & 20.9 \\
ast-ps & 22.3 & 22.7 & 23.0 & 4.8 & 4.8 & 5.0 \\
ast-ur & 31.4 & 33.0 & 33.4 & 10.5 & 11.6 & 11.8 \\
ast-zu & 13.7 & 25.3 & 27.9 & 1.8 & 4.9 & 5.6 \\
hr-af & 50.8 & 50.7 & 50.9 & 23.4 & 23.3 & 23.2 \\
hr-ast & 47.3 & 48.5 & 48.3 & 20.6 & 20.1 & 20.0 \\
hr-ps & 24.0 & 24.1 & 24.4 & 5.6 & 5.4 & 5.4 \\
hr-ur & 35.2 & 35.4 & 35.7 & 13.3 & 13.4 & 13.3 \\
hr-zu & 21.7 & 28.9 & 30.4 & 3.2 & 6.0 & 6.3 \\
ps-af & 39.0 & 39.2 & 39.2 & 12.0 & 12.2 & 12.3 \\
\textit{ps-ast} & \textit{29.9} & \textit{34.8} & \textit{35.0} & \textit{6.0} & \textit{9.3} & \textit{10.0} \\
ps-hr & 35.3 & 35.7 & 35.8 & 9.4 & 9.8 & 9.8 \\
ps-ur & 31.5 & 31.5 & 31.8 & 10.2 & 10.4 & 10.4 \\
ps-zu & 15.8 & 21.1 & 23.2 & 1.0 & 2.3 & 3.0 \\
ur-af & 42.6 & 42.9 & 43.1 & 15.1 & 15.1 & 15.1 \\
ur-ast & 33.7 & 38.5 & 38.3 & 8.3 & 12.1 & 12.1 \\
ur-hr & 40.4 & 40.4 & 40.6 & 13.4 & 13.3 & 13.2 \\
ur-ps & 23.5 & 23.8 & 23.9 & 5.1 & 5.1 & 5.2 \\
ur-zu & 11.6 & 19.5 & 20.6 & 0.6 & 2.1 & 2.6 \\
zu-af & 33.8 & 35.5 & 35.6 & 8.9 & 11.1 & 11.2 \\
zu-ast & 26.8 & 31.4 & 32.0 & 4.9 & 7.5 & 8.6 \\
zu-hr & 29.1 & 31.4 & 31.8 & 5.5 & 7.4 & 7.7 \\
zu-ps & 15.1 & 18.2 & 18.1 & 1.4 & 2.6 & 2.4 \\
zu-ur & 22.0 & 25.1 & 25.2 & 3.4 & 5.2 & 5.2 \\
\midrule
avg (X-branch) & 32.0 & 34.2 & 34.6 & 10.2 & 11.1 & 11.2 \\
\midrule\midrule
de-en & 62.7 & 62.3 & 62.2 & 38.3 & 37.4 & 37.3 \\
en-de & 59.3 & 58.9 & 58.8 & 33.7 & 33.2 & 32.9 \\
en-fr & 64.8 & 64.2 & 64.1 & 43.4 & 41.9 & 41.8 \\
fr-en & 63.2 & 63.0 & 62.7 & 39.4 & 39.0 & 38.6 \\
\midrule
avg (high-res) & 62.5 & 62.1 & 62.0 & 38.7 & 37.9 & 37.6 \\
\midrule\midrule
avg (all) & 41.4 & 42.5 & 42.7 & 18.1 & 18.4 & 18.4 \\
\bottomrule
\end{tabular}
\end{adjustbox}
\caption{Full results for SMaLL-100. Averages exclude ps-ast translation direction for comparability to M2M-100.}
\label{all-small}
\end{table*}

\clearpage

\section{Languages}\label{sec:appendix:langs}
\begin{table*}[hbt!]
\centering
\begin{tabular}{ll}
\toprule
language code & language\\
\midrule
af & Afrikaans \\
ar & Arabic \\
ast & Asturian \\
be & Belarusian \\
bn & Bengali \\
cs & Czech \\
de & German \\
el & Greek\\
en & English\\
fr & French\\
hi & Hindi\\
hr & Croatian\\
hu & Hungarian\\
hy & Armenian\\
it & Italian\\
mr & Marathi\\
nl & Dutch; Flemish\\
ps & Pushto; Pashto\\
ro & Romanian; Moldavian; Moldovan \\
ru & Russian\\
sk & Slovak \\
sr & Serbian \\
sw & Swahili \\
tr & Turkish \\
uk & Ukrainian \\
ur & Urdu\\
zu & Zulu \\

\bottomrule
\end{tabular}
\caption{List of languages in our experiments, sorted by ISO 639-1 language code.}
\end{table*}

\vfill

\clearpage

\section{LLM Off-Target Analysis for English--French}\label{sec:llm-results-en-fr}

\begin{figure}[hbt!]
\centering
\resizebox{0.5\columnwidth}{!}{%
\begin{tikzpicture}
\begin{axis}[
    height=6.5cm,
    width=9cm,
    axis lines = left,
    xlabel = $\lambda_{\text{lang}}$,
    ylabel = {off-target rate},
    ymin=0, ymax=4,
    xtick={0,0.1,0.3,0.5,0.7,0.9},
    yticklabel={\pgfmathprintnumber\tick\%},
    legend pos=north east,
    legend cell align={left},
]
\addplot[
    color={rgb,255:red,153; green,169; blue,219},
    mark=*,
    line width=1.5pt,
    dashed,
    ]
    coordinates {
    (0,2.08) (0.1,1.78) (0.3,1.68) (0.5,1.48) (0.7,1.48) (0.9,0.99)
    };
\addlegendentry{Llama 7B 0-shot}
\addplot[
    color={rgb,255:red,153; green,169; blue,219},
    mark=*,
    line width=1.5pt,
    ]
    coordinates {
    (0,1.78) (0.1,1.28) (0.3,1.09) (0.5,1.09) (0.7,0.99) (0.9,0.59)
    };
\addlegendentry{Llama 7B 1-shot}
\addplot[
    color={rgb,255:red,0; green,40; blue,165},
    mark=*,
    dashed,
    line width=1.5pt,
    ]
    coordinates {
    (0,2.17) (0.1,1.98) (0.3,1.48) (0.5,1.28) (0.7,0.89) (0.9,0.49)
    };
\addlegendentry{Llama 13B 0-shot}
\addplot[
    color={rgb,255:red,0; green,40; blue,165},
    mark=*,
    line width=1.5pt,
    ]
    coordinates {
    (0,2.17) (0.1,1.88) (0.3,0.79) (0.5,0.49) (0.7,0.30) (0.9,0.40)
    };
\addlegendentry{Llama 13B 1-shot}
\end{axis}
\end{tikzpicture}
}
\caption{Off-target translation rate for Llama~2 Chat models when translating the English Flores-101 devtest set into French.
As with German~(Figure~\ref{fig:llm-en-de}), language-contrastive decoding tends to reduce off-target translation as~$\lambda_{\text{lang}}$ is increased.
}
\label{fig:llm-en-fr}
\end{figure}

\vfill

\section{LLM Automatic Evaluation Results}\label{sec:llm-results}

\begin{table*}[hbt!]
\centering
\footnotesize
\begin{adjustbox}{width=1\textwidth}
\begin{tabular}{lcccccccccccc}
\toprule
en-de & \multicolumn{6}{c}{chrF2} & \multicolumn{6}{c}{spBLEU}\\
& baseline & $\lambda_\text{lang}=0.1$ & 0.3 & 0.5 & 0.7 & 0.9 & baseline & $\lambda_\text{lang}=0.1$ & 0.3 & 0.5 & 0.7 & 0.9\\
\cmidrule(r){2-7} \cmidrule(l){8-13}
Llama 7B 0-shot & 50.0 & 49.9 & 50.2 & 50.3 & 49.9 & 49.4 & 23.8 & 23.7 & 23.8 & 23.7 & 23.3 & 22.3 \\
Llama 7B 1-shot & 50.5 & 50.9 & 51.1 & 51.4 & 50.9 & 49.7 & 24.4 & 24.7 & 24.8 & 25.1 & 24.3 & 22.6 \\
Llama 13B 0-shot & 54.2 & 54.5 & 54.5 & 54.7 & 54.3 & 53.3 & 29.1 & 29.4 & 29.3 & 29.3 & 29.0 & 27.8 \\
Llama 13B 1-shot & 54.4 & 54.5 & 54.7 & 55.1 & 54.9 & 53.7 & 29.4 & 29.5 & 29.7 & 29.9 & 29.5 & 27.4 \\
\midrule
\textit{Average} & \textit{52.3} & \textit{52.5} & \textit{52.6} & \textit{52.9} & \textit{52.5} & \textit{51.5} & \textit{26.7} & \textit{26.8} & \textit{26.9} & \textit{27.0} & \textit{26.5} & \textit{25.0}\\
\bottomrule
\end{tabular}
\end{adjustbox}
\caption{English--German: Automatic evaluation of LLM-based translation on the Flores-101 devtest set.
The scores tend to increase with smaller values of $\lambda_{\text{lang}}$, but decline with larger values.
}
\label{llm-en-de-results}
\end{table*}

\begin{table*}[hbt!]
\centering
\footnotesize
\begin{adjustbox}{width=1\textwidth}
\begin{tabular}{lcccccccccccc}
\toprule
en-fr & \multicolumn{6}{c}{chrF2} & \multicolumn{6}{c}{spBLEU}\\
& baseline & $\lambda_\text{lang}=0.1$ & 0.3 & 0.5 & 0.7 & 0.9 & baseline & $\lambda_\text{lang}=0.1$ & 0.3 & 0.5 & 0.7 & 0.9\\
\cmidrule(r){2-7} \cmidrule(l){8-13}
Llama 7B 0-shot & 58.3 & 58.7 & 58.8 & 58.6 & 58.1 & 57.2 & 35.2 & 35.6 & 35.7 & 35.5 & 34.9 & 33.5 \\
Llama 7B 1-shot & 58.4 & 58.7 & 58.7 & 58.4 & 58.0 & 56.7 & 35.8 & 36.2 & 36.1 & 35.7 & 35.1 & 33.2 \\
Llama 13B 0-shot & 62.4 & 62.5 & 62.6 & 62.6 & 62.6 & 62.0 & 40.6 & 40.6 & 40.8 & 40.8 & 40.6 & 39.7 \\
Llama 13B 1-shot & 62.1 & 62.2 & 62.6 & 62.6 & 62.6 & 61.7 & 40.6 & 40.7 & 41.0 & 41.2 & 41.1 & 39.8 \\
\midrule
\textit{Average} & \textit{60.3} & \textit{60.5} & \textit{60.7} & \textit{60.6} & \textit{60.3} & \textit{59.4} & \textit{38.0} & \textit{38.3} & \textit{38.4} & \textit{38.3} & \textit{37.9} & \textit{36.6}\\
\bottomrule
\end{tabular}
\end{adjustbox}
\caption{English--French: Automatic evaluation of LLM-based translation on the Flores-101 devtest set, showing patterns similar to English--German.
}
\label{llm-en-fr-results}
\end{table*}

\vfill

\clearpage

\section{LLM Implementation Details}\label{sec:llm-implementation-details}
Our input to Llama consists of a {\color{gray}system prompt} and an {\color{blue} instruction}. We force-decode the {\color{orange}prefix of the assistant response} to make sure that the next generated line is the actual translation and not a prologue by the assistant.

\subsection*{Zero-shot}
\small

\texttt{<s>[INST] {\color{darkgray} \textless{}\textless{}SYS\textgreater{}\textgreater{}} \\
\noindent{\color{darkgray}You are a machine translation system that translates sentences from English to German. You just respond with the translation, without any additional comments.} \\
\noindent{\color{darkgray}\textless{}\textless{}/SYS\textgreater{}\textgreater{}}[INST] {\color{blue}"We now have 4-month-old mice that are non-diabetic that used to be diabetic," he added.}} \\
\\
\noindent{}\texttt{{\color{blue}Translate to German} [/INST]{\color{orange}Sure, here's the translation:}}

\subsection*{One-shot}
\small

\noindent{}\texttt{<s>[INST] {\color{darkgray}\textless{}\textless{}SYS\textgreater{}\textgreater{}} \\
\noindent{\color{darkgray}You are a machine translation system that translates sentences from English to German. You just respond with the translation, without any additional comments.}} \\
\\
\noindent{}\texttt{{\color{darkgray}Example instruction:}} \\
\\
\noindent{}\texttt{{\color{darkgray}On Monday, scientists from the Stanford University School of Medicine announced the invention of a new diagnostic tool that can sort cells by type: a tiny printable chip that can be manufactured using standard inkjet printers for possibly about one U.S. cent each.
}} \\
\noindent{}\texttt{{\color{darkgray}Translate to German}} \\
\\
\noindent{}\texttt{{\color{darkgray}Example response:}} \\
\\
\noindent{}\texttt{{\color{darkgray}Sure, here's the translation:}} \\
\noindent{}\texttt{{\color{darkgray}Am Montag haben die Wisenschaftler der Stanford University School of Medicine die Erfindung eines neuen Diagnosetools bekanntgegeben, mit dem Zellen nach ihrem Typ sortiert werden können: ein winziger, ausdruckbarer Chip, der für jeweils etwa einen US-Cent mit Standard-Tintenstrahldruckern hergestellt werden kann.}} \\
\noindent{}\texttt{{\color{darkgray}\textless{}\textless{}/SYS\textgreater{}\textgreater{}}[INST] {\color{blue}"We now have 4-month-old mice that are non-diabetic that used to be diabetic," he added.}} \\
\\
\noindent{}\texttt{{\color{blue}Translate to German} [/INST]{\color{orange}Sure, here's the translation:}}

\normalsize

\vfill

\end{document}